\documentclass[runningheads]{llncs}
\usepackage{amsmath,amssymb,amsfonts}
\usepackage{graphicx}
\usepackage{textcomp}
\usepackage{xcolor}
\usepackage{hyperref}
\usepackage{booktabs}
\usepackage{subcaption}
\usepackage{cleveref}
\usepackage{bbm}
\usepackage{multirow}
\usepackage{biblatex}
\usepackage{enumitem}
\setlist[itemize]{topsep=4pt, itemsep=2pt, parsep=0pt}
\newcommand{\numimages}{12634}
\newcommand{\numsegments}{120044}
\newcommand{\numcategories}{239}

\usepackage{tikz}
\usetikzlibrary{arrows.meta, fit, backgrounds, calc, positioning}

\definecolor{procfill}{RGB}{235,242,250}   
\definecolor{boxedge} {RGB}{80, 100, 130}  

\begin{document}


\title{Efficient Image Annotation via Semi-Supervised Object Segmentation with Label Propagation}
\titlerunning{Efficient Annotation with Label Propagation}
\author{
Vitalii Tutevych$^*$ \and
Raphael Memmesheimer$^*$ \and
Luca Eichler \and
Dmytro Pavlichenko \and
Fynn Schilke \and
Rodja Krudewig \and
Sven Behnke
}
\renewcommand{\thefootnote}{\fnsymbol{footnote}}
\footnotetext[1]{These authors contributed equally to this work.}
\authorrunning{V. Tutevych \and R. Memmesheimer et al.}

\institute{
Autonomous Intelligent Systems Group, University of Bonn, Germany \\
\email{memmesheimer@ais.uni-bonn.de}
}

\maketitle

\begin{abstract}
Reliable object perception is necessary for general-purpose service robots. Open-vocabulary detectors struggle to generalize beyond a few classes and fully supervised training of object detectors requires time-intensive annotations.
We present a semi-supervised label propagation approach for household object segmentation.
A segment proposer generates class-agnostic masks, and an ensemble of Hopfield networks assigns labels by learning representative embeddings in complementary foundation model embedding spaces (CLIP, ViT, Theia).
Our approach scales to 50 object classes with limited annotation overhead and can automatically label 60\% of the data in a RoboCup@Home setting, where preparation time is severely constrained.
Dataset and code are publicly available at \url{https://github.com/ais-bonn/label_propagation}.
\end{abstract}


\section{Introduction}

Robots interacting with objects in an environment require them to perceive these objects. 
The creation of accurately labeled object segmentation datasets is a time-intensive and exhaustive effort. 
This is especially a problem in time-constrained settings with many class instances e.g. for robot competition settings like RoboCup@Home~\cite{matamoros2018robocup}.
RoboCup@Home is a robot competition where robots perform household tasks in realistic apartments autonomously. 
The difficulty for perception lies in the fact that upon arrival teams are provided with a list of ca. 50 different classes of objects that need to be recognized and manipulated. 
There is limited preparation (commonly one or two days) to record training data and train a model for the robot to perform the tasks. 
While recent advances in contrastive language image pairing~\cite{radford2021learning} lay the foundation for open-vocabulary object detection~\cite{liu2024grounding} and segmentation approaches~\cite{zhou2022detecting}. Furthermore, foundational models for segmentation have recently been very influential~\cite{kirillov2023segment}. In the case of constrained object sets, we found that these open vocabulary-driven approaches tend to perform well with lower amounts of classes while not yet generalizing to the larger amount of classes. 

In this paper, we present a two-stage process. First, segment proposals are trained on a large object segmentation dataset gathered over competition attendances and utilized to estimate object segments. Second, a set of reference images per object class is used to find a metastable representation of each class which is compared to the input embedding in Foundation Model representation space for association. This approach massively supports human labelers by suggesting 60\% of the labels. Finally, a fully supervised model is trained that performs well on many object classes, including some that are very similar.  This focused propagation drastically reduces the influence of outliers and enhances the accuracy of label assignments.

We summarize the contributions of this paper as follows:
\begin{itemize}
    \item We propose a semi-automatic approach that guides the labeling process by estimating general object segments and propagating reference labels for classes among the whole dataset. 
    \item With the approach, we provide the code publicly and dataset that we utilized for training the general object segmentation model and evaluated the semi-automatic label propagation approach. 
\end{itemize}

\section{Related Work}

Currently, segmentation models following a supervised training~\cite{carion2020end,jocher2022ultralytics, li2023mask} scheme still define the standard for robot competitions~\cite{memmesheimer2024robocup, memmesheimeradaptive, memmesheimernimbro} where robots are required to recognize around 50 different objects in household environments. Open vocabulary approaches~\cite{zhou2022detecting,  liu2024grounding} applied in these competitions~\cite{memmesheimer2024robocup} have been shown to perform well with lower class amounts. In this paper, we propose a model that aims to close the gap and support the labeling process by proposing masks that relate to objects commonly utilized in these competitions and semi-automatically suggesting labels from a few representatives by calculating their similarity to the proposed masks. This is achieved by the utilization of models trained with Contrastive Language-Image Pre-Training (CLIP)~\cite{radford2021learning}, which provides a powerful multimodal embedding space.

\paragraph*{Semantic Segmentation}
Traditional semantic segmentation approaches rely on manually annotated segment annotations and are trained in a supervised context. Common representatives from this domain are YOLO~\cite{jocher2022ultralytics} which is widely spread in robotic competition domains. YOLO mainly dominates the segmentation by the ease of use for both inference and training.
DETR~\cite{carion2020end} reformulated the object detection problem as a transformer-based detection that maps, while MaskDINO~\cite{li2023mask} extended this approach to a pixel-wise semantic segmentation. Both models improved the state-of-the-art detection and segmentation results. Recently open-vocabulary approaches like Detic~\cite{zhou2022detecting} Grounding-DINO~\cite{liu2024grounding} enabled promptable semantic segmentation.
\citeauthor{gouda2024learning}~\cite{gouda2024learning} integrate a zero-shot segmentation model (like SAM) with a Centroid Triplet Loss (CTL) trained backbone to match query images of objects. This approach is similar to our approach, but we in addition use a Segment Proposer to generate class-agnostic masks and associate labels from reference sets with an ensemble of Hopfield networks.
\citeauthor{patzold2025leveraging}~\cite{patzold2025leveraging} guide open vocabulary detectors generated by input prompts from visual language models.

\paragraph*{Datasets}

\citeauthor{massouh2019robocup}~\cite{massouh2019robocup} proposed a large-scale object recognition consisting of household objects. This dataset, collected by downloading them from web image searches, consists of 196,000 images in the training set and spans 180 classes. The dataset focuses on class annotations per image and therefore is not suitable for object detection and object segmentation approaches. In contrast to this dataset, our dataset provides segmentation annotations, which allows for further robotic downstream tasks like mobile manipulation.
\citeauthor{7wxn-n828-20}~\cite{7wxn-n828-20} provide a dataset of annotated images targeting RoboCup@Home objects. Their dataset consists of two sets with a total of 554 annotated images and 2765 labeled bounding boxes for objects. Our dataset  also consists of pixel-level segmentations and includes more images as well as labeled instances.
\citeauthor{novkovic2019clubs}~\cite{novkovic2019clubs} proposed a dataset containing multiple object scenes with single objects and box scenes with cluttered household object arrangements. They collected multimodal data from four different sensors.
\citeauthor{tyree20226}~\cite{tyree20226} present the HOPE dataset, which consists of 2088 images from 10 different scenes labeled on an object segment and pose level.
Our dataset has the benefit of capturing multiple local household objects captured in multiple countries on a semantic segmentation level.
\begin{figure}[t]
    \centering
    \scalebox{.7}{\input{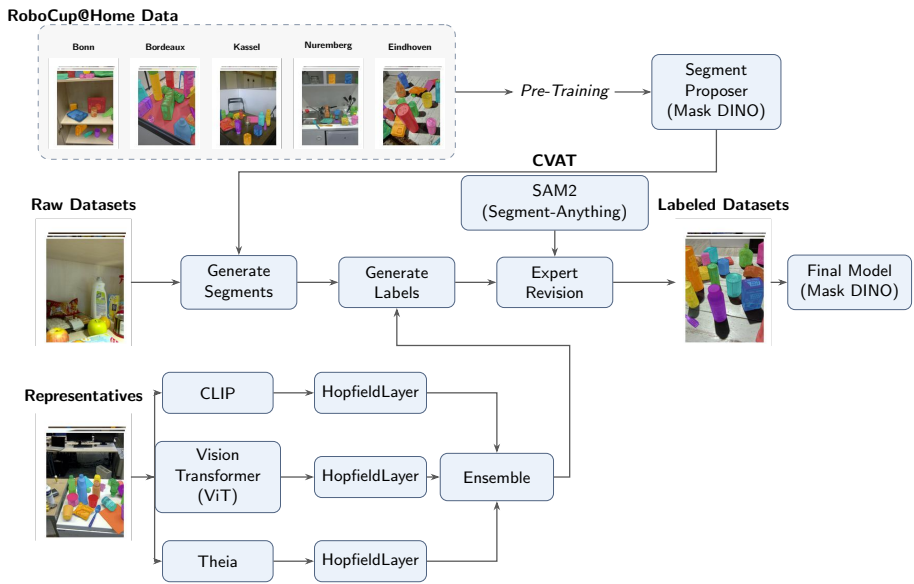}}
    
    \vspace{-0.2cm}
    \caption{Labeling and training pipeline. To train the final object detector, we first utilize a Segment Proposer model pre-trained on the data from previous competitions. After initial pre-labeling, a human expert verifies and corrects the generated labels for the representative dataset. The labeler is then used with a general object detector to generate improved annotations for the final detector.}
    \label{fig:pipeline_architecture}
    \vspace{-0.4cm}
\end{figure}
\section{Approach}

We first train a Segment Proposer to generate class-agnostic masks, and then train an ensemble of Hopfield networks to assign labels by learning representative embeddings in complementary Foundation Model embedding spaces (CLIP, ViT, Theia).
The approach is summarized in \textit{Figure \ref{fig:pipeline_architecture}}.
In the following, we describe the data collection process, the Segment Proposer, the Hopfield Labeler, and the final detector training.

\begin{figure}[t]
    \centering
    \scalebox{0.4}{
    \begin{tikzpicture}
      \node[anchor=south west, inner sep=0] (image) at (0,0)
        {\includegraphics[width=12cm]{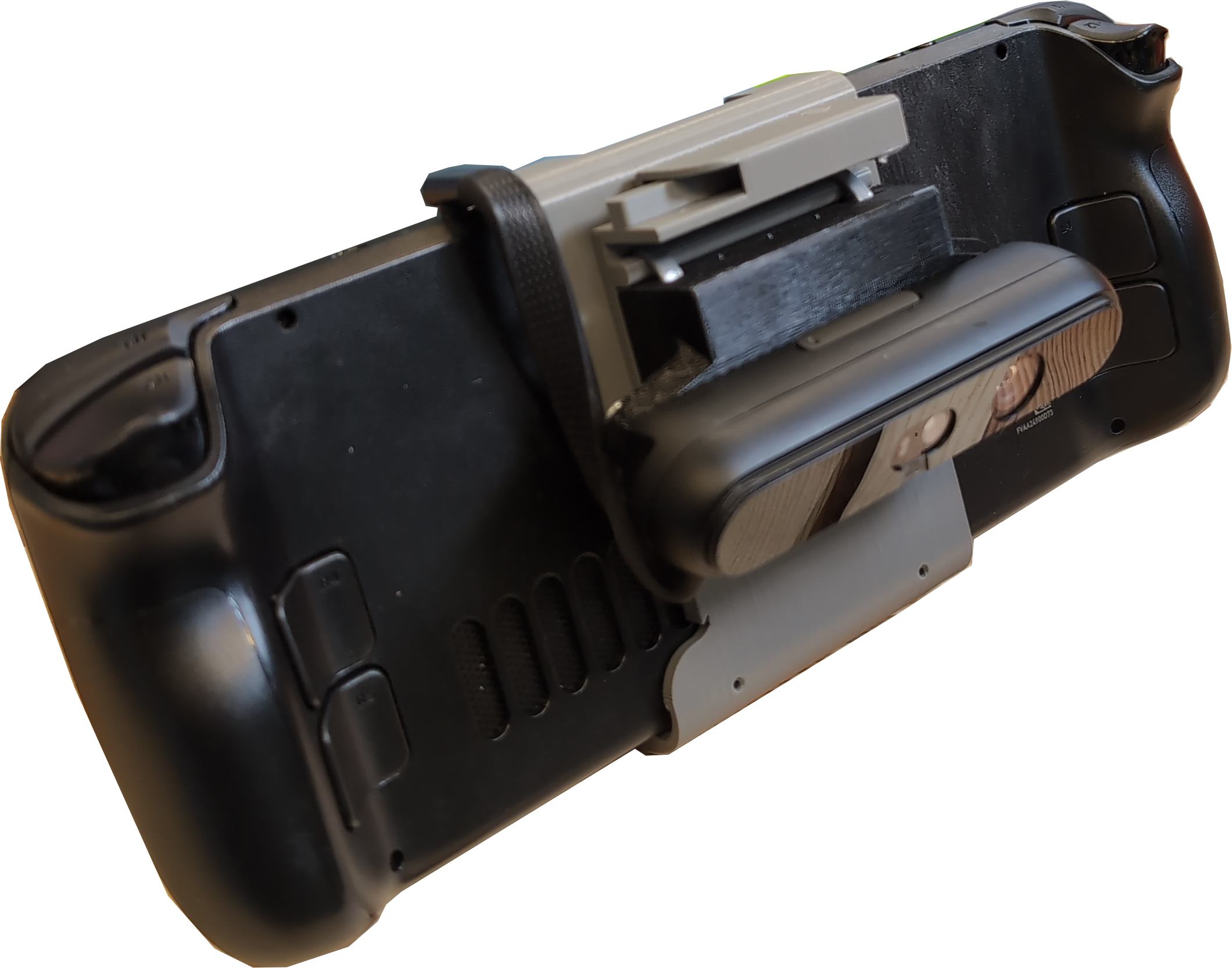}};
    
    
      \coordinate (gemini)       at ($(image.south west)!1.6!(image.south east)!0.560!(image.north west)$);
      \coordinate (geminiLabel)  at ($(image.south west) + (8cm, 3cm)$);
      \draw[white, line width=1.2pt, -{Stealth[length=8pt]}]
        (geminiLabel) -- (gemini);
      \node[white, anchor=west, font=\sffamily\bfseries\large,
            fill=black!70, rounded corners=3pt, inner sep=4pt]
        at (geminiLabel) {Orbbec Gemini 2};
    
      \coordinate (deck)       at ($(image.south west)!0.226!(image.south east)!0.310!(image.north west)$);
      \coordinate (deckLabel)  at ($(image.south west) + (0cm, 2.0cm)$);
      \draw[white, line width=1.2pt, -{Stealth[length=8pt]}]
        (deckLabel) -- (deck);
      \node[white, anchor=west, font=\sffamily\bfseries\large,
            fill=black!70, rounded corners=3pt, inner sep=4pt]
        at (deckLabel) {Valve Steam Deck};
    
    \end{tikzpicture}
    \vspace{-0.2cm}
    }
    \caption{Data-recording setup: We utilized an Orbbec Gemini 2 connected to a Valve Steam Deck supported by a guided data-recording procedure.}
    \vspace{-0.2cm}
    \label{fig:data_recording}
\end{figure}
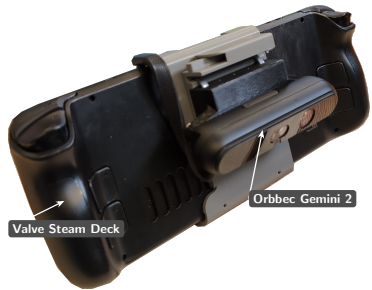

\subsection{Data Collection}
\label{ssec:data_collection}

We collect two types of data at each competition venue using the setup shown in \Cref{fig:data_recording}.
A \emph{representative dataset} contains around 50 objects with 4-8 images per object, captured from four cardinal directions at two distances, yielding 1600–2000 instances per venue.
The \emph{Training/validation datasets} consist of 144 training and 16 validation images per set, with 2–3 sets per venue, featuring object overlaps, occlusions, and challenging lighting.

\begin{figure}[t]
    \centering
    \begin{subfigure}[b]{0.49\linewidth}
        \includegraphics[width=\linewidth]{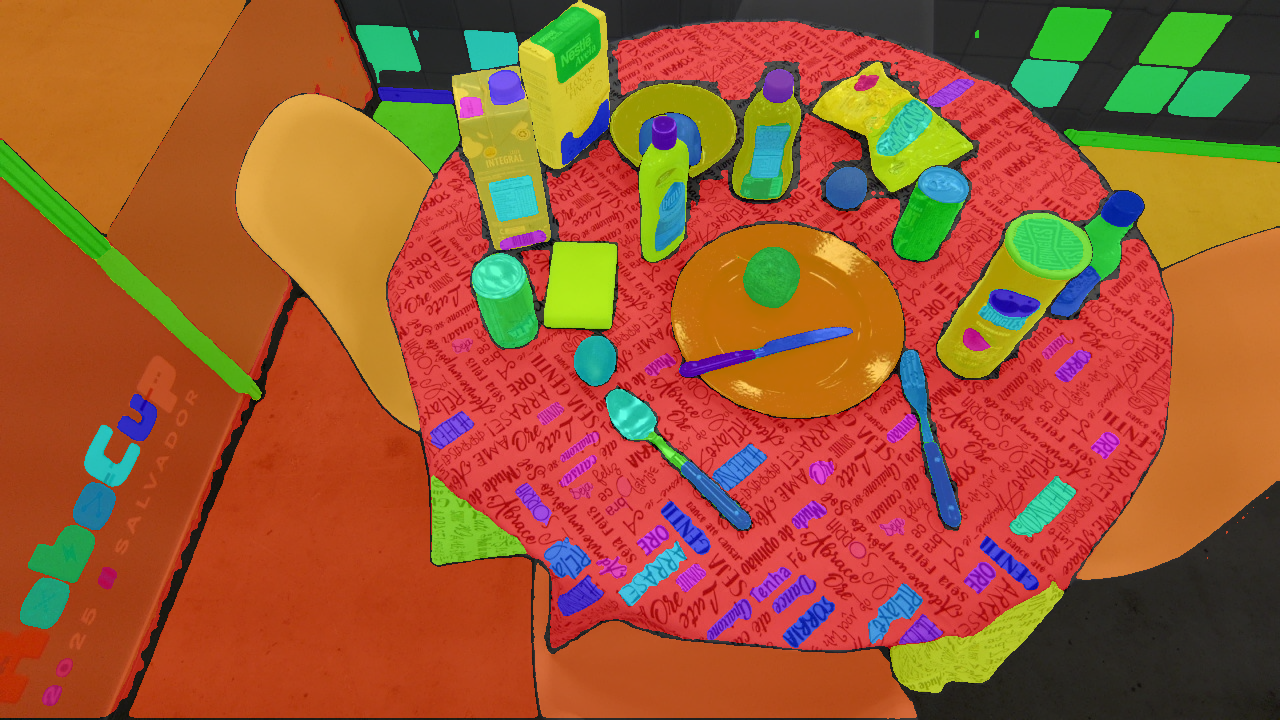}
        \caption{Segment Anything}
        \label{fig:segment_anything}
    \end{subfigure}
    \begin{subfigure}[b]{0.49\linewidth}
        \includegraphics[width=\linewidth]{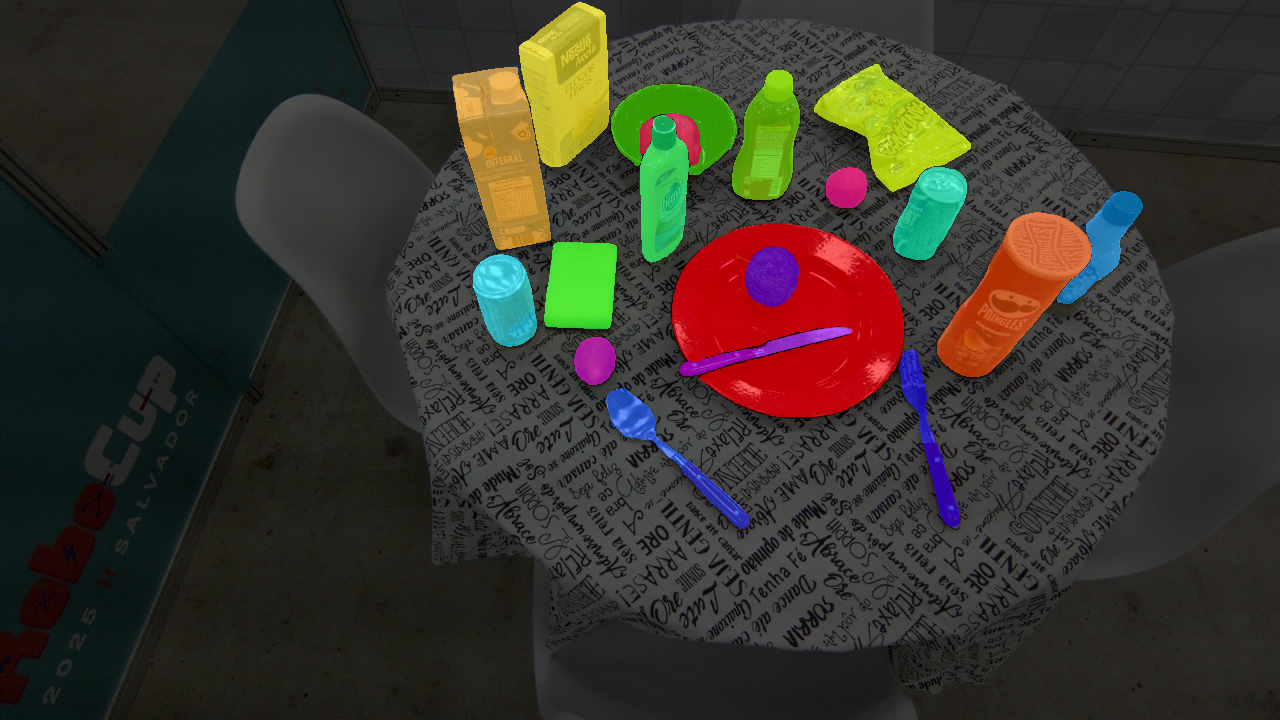}
        \caption{Segment Proposer (Ours)}
        \label{fig:segment_proposer}
    \end{subfigure}
    \vspace{-0.2cm}
    \caption{Qualitative example of the Segment Anything model in comparison to our proposed Segment Proposer for household objects.}
    \label{fig:segment_proposer_comparison}
 \end{figure}

\subsection{Segment Proposer}

We introduce a segment proposer model that automates object segmentation in images, eliminating the need for manual mask annotation by human annotators. Unlike the guided Segment Anything approach—which requires manual selection of positive/negative points—our custom-trained model specializes in household objects and proves more efficient.
Trained on annotated data from the Bordeaux, Eindhoven, and Kassel datasets, the model streamlined labeling for the Nuernberg and Salvador sets. \Cref{fig:segment_proposer_comparison} compares our approach to Segment Anything, highlighting its simplicity and reduced computational overhead. While coupling Segment Anything with filtering/detection models is possible, it introduces unnecessary complexity.

\subsection{Hopfield Labeler}

\begin{figure}[t]
    \centering
    \includegraphics[width=0.8\linewidth]{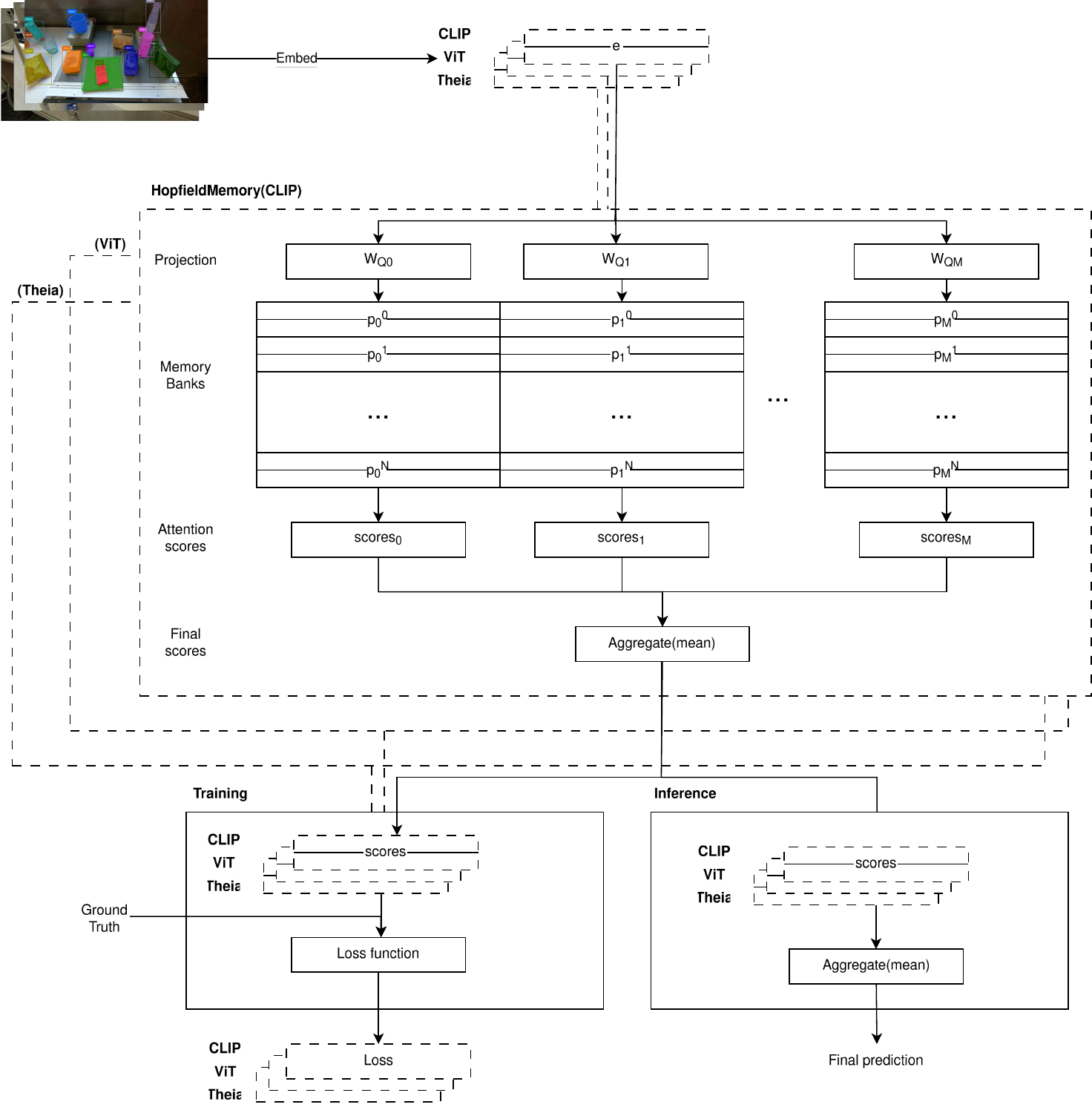}
    \vspace{-0.2cm}
    \caption{Labeler architecture. One Hopfield head is trained per foundation model; outputs are combined by mean aggregation at inference time.}
    \label{fig:labeler_architecture}
\end{figure}

The Labeler (\Cref{fig:labeler_architecture}) assigns a class label to each proposed mask by comparing its Foundation Model embedding against a set of learned \emph{representative} embeddings — one per class.
We decouple the Labeler from the Segment Proposer: labels are inferred from small representative crops; the Labeler is never fine-tuned on full detection datasets.

\textbf{Foundation Model embeddings.}
Each annotated patch is encoded by a frozen Foundation Model.
We use OpenAI CLIP, Vision Transformer (ViT), and Theia~\cite{shang2024theia}.
For transformer-based models the \texttt{[CLS]} token serves as the patch representation; Theia provides a direct bottleneck vector.

\textbf{Hopfield Memory.}
We store one learned representative embedding per class in a Hopfield Memory~\cite{ramsauer2020hopfield}.
At inference the attention score between a query embedding $R$ and the memory matrix $Y$ is:
\begin{equation}
    \mathrm{scores} = \mathrm{softmax}\!\left(\beta\, R W_Q W_K^\top Y^\top\right).
\end{equation}
Training minimizes the MSE between predicted scores and one-hot class targets:
\begin{equation}
    \mathcal{L} = \frac{1}{N}\sum_{i=1}^{N}\bigl(\mathrm{scores}(x_i) - \mathbbm{1}_{y_i}\bigr)^2.
\end{equation}
This is equivalent to simultaneously pulling each class representative toward its positive examples while pushing all other representatives away — a supervised analogue of contrastive learning.

\textbf{Bank subdivision and regularization.}
To improve robustness we split the Hopfield Memory into $m$ banks, each operating in a lower-dimensional projection space similar to multi-head attention.
Final scores are the mean across banks.
Two regularizes enforce complementarity: an \emph{intra-bank} term clusters representatives within each bank, and an \emph{inter-bank} term pushes the same-class representatives of different banks apart.
Empirically, this subdivision together with the regularizes consistently improves recognition accuracy.

\textbf{Ensemble.}
One Hopfield head is trained independently per Foundation Model embedding space.
At inference the predictions of all three heads are combined by mean aggregation. We neither train nor finetune the Ensemble model in any way after training of the individual heads completes.

\subsection{Final Detector Training}

The Segment Proposer and Hopfield Ensemble generate annotations for the training/validation sets.
A human expert reviews and corrects these annotations in CVAT~\cite{cvat_ai_corporation_2023_7863887} before training the final detector (MaskDinoV2 via Detectron2) used on the robot.
We study the impact of skipping this review step in \textbf{Q4}.

\section{Experiments}

We address four questions:
\textbf{Q1} Does learning representative embeddings via Hopfield Memory outperform fixed cosine-similarity retrieval?
\textbf{Q2} Is an ensemble of independently trained heads more accurate than any single head?
\textbf{Q3} Does the full pipeline reduce annotation effort in practice?
\textbf{Q4} Can pipeline-generated annotations replace expert annotations for detector training?
\begin{figure*}[t]
    \centering
    \begin{subfigure}[b]{0.327\textwidth}
        \includegraphics[width=\linewidth]{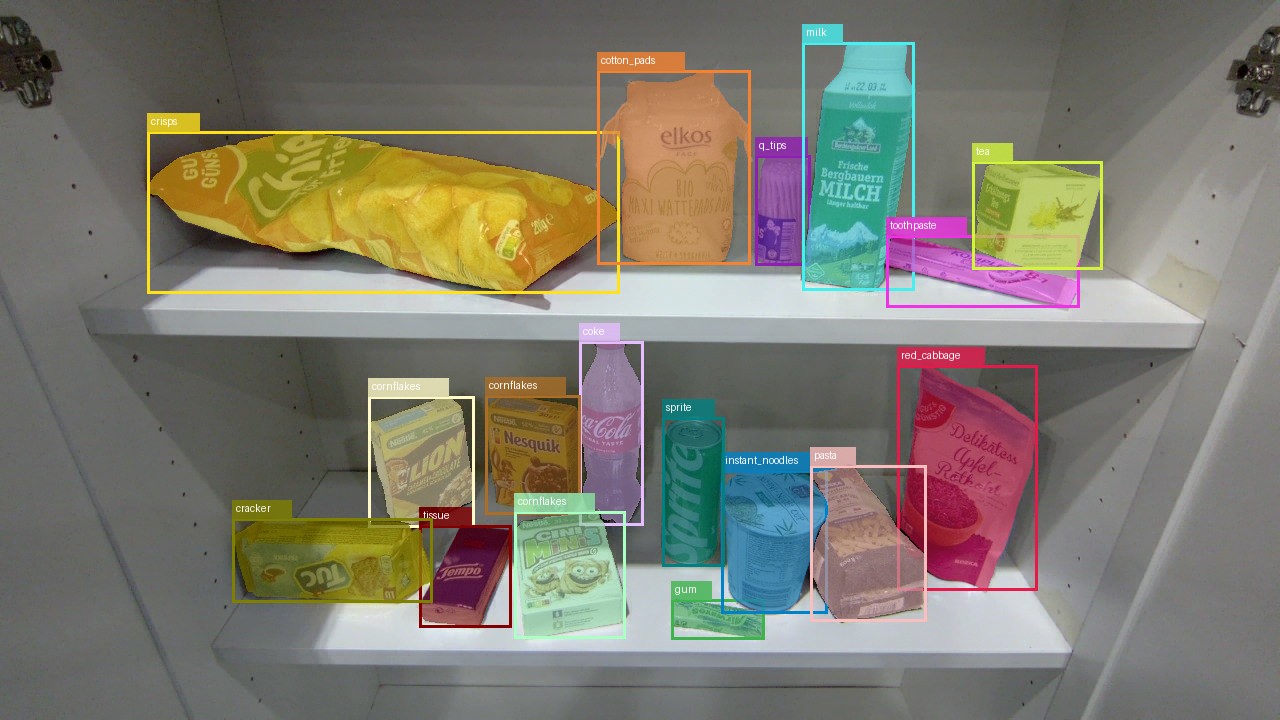}
        \caption{Nuernberg}
        \label{fig:example_nuernberg}
    \end{subfigure}
    \begin{subfigure}[b]{0.327\textwidth}
        \includegraphics[width=\linewidth]{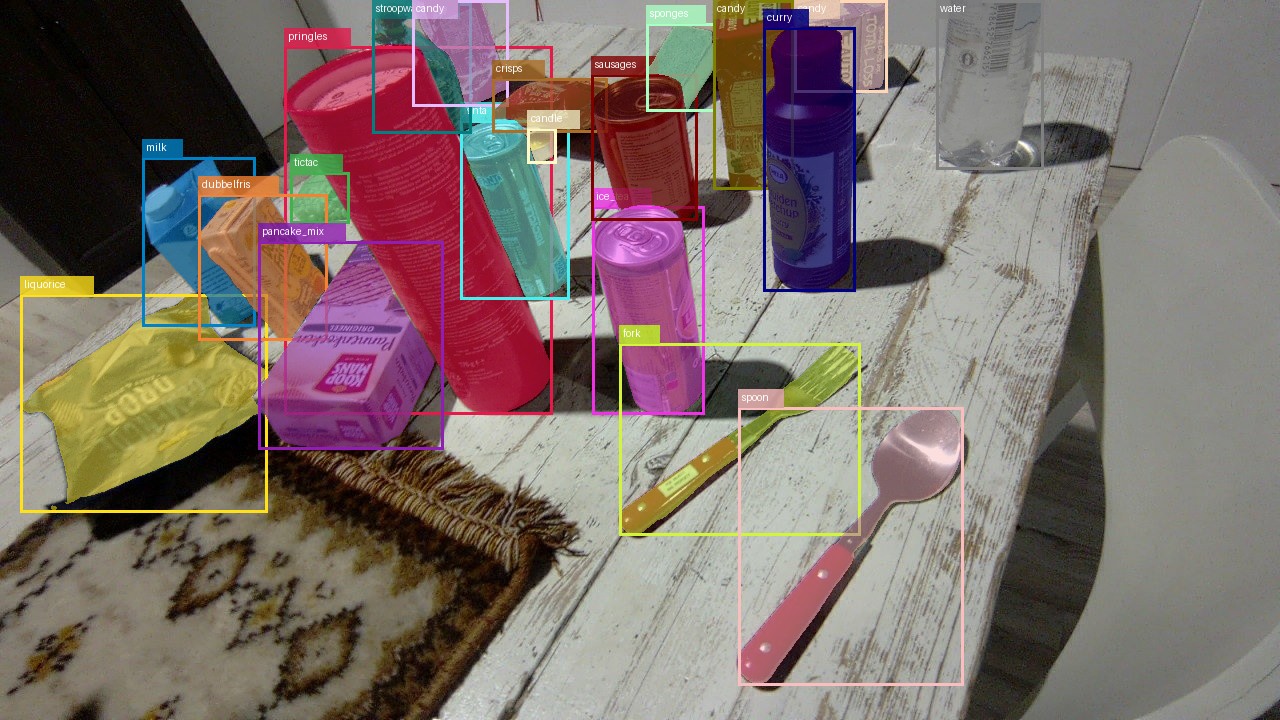}
        \caption{Eindhoven}
        \label{fig:example_eindhoven}
    \end{subfigure}
    \begin{subfigure}[b]{0.327\textwidth}
        \includegraphics[width=\linewidth]{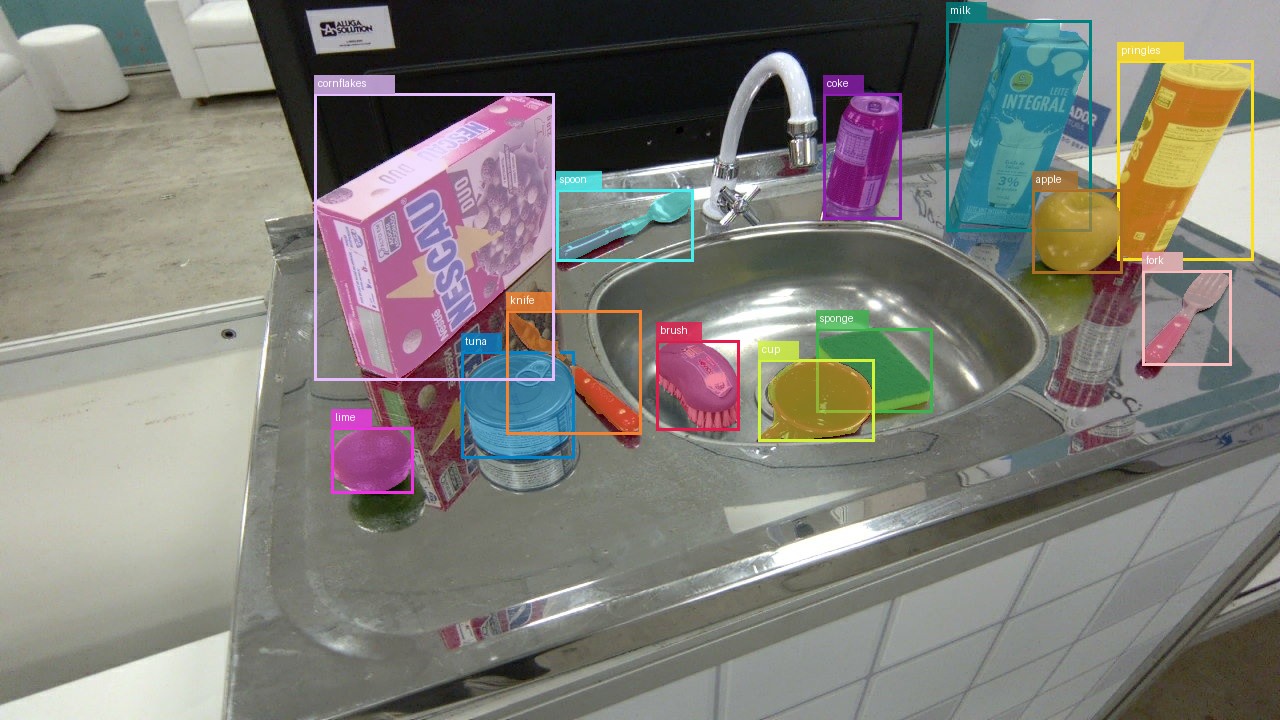}
        \caption{Salvador}
        \label{fig:example_salvador}
    \end{subfigure}
    \vspace*{-0.3cm}
    \caption{Labeled example images from different competition venues.}
    \vspace*{-0.3cm}
    \label{fig:example_images}
\end{figure*}

\subsection{Dataset}

Following the collection procedure of \Cref{ssec:data_collection} we assembled a dataset of \numsegments{} segmentation annotations across \numcategories{} object categories in \numimages{} images.
Data were collected at seven competition venues across four countries France, Germany, the Netherlands, and Brazil, reflecting diverse local household products.
Example annotations are shown in \Cref{fig:example_images}.

\subsection{Training and Deployment Details}
Training uses a PC with a 24\,GB NVIDIA GeForce RTX~4090.
The Labeler requires $\sim$10\,GB VRAM and trains in 1.5\,h; the Final Detector (MaskDinoV2) uses the full card and trains in 30\,min with 2400 iterations, batch size 4.
Full machine annotation of one dataset with ca. 160 images takes 2–3 minutes.
Images are resized to $224{\times}224$ and augmented with random crops, flips, affine transforms, colour jitter, grayscale, and Gaussian blur during Labeler training.
Key hyperparameters: learning rate $0.001$, 20 epochs, batch size 16, $\lambda_{\text{inter}}=0.01$, $\lambda_{\text{intra}}=0.1$.

\subsection{Evaluation and Analysis}
\textbf{Evaluation Setup}
The trained models are to be evaluated using the metrics of Mask Precision, Mask Recall and Mask Mean Average Precision (mAP). We use FiftyOne for model evaluation on gathered data and report the metrics provided by the library.

\textbf{Q1} For the first experiment we compare the respective Hopfield heads with simply storing a number of representative embeddings and comparing them to the query embedding in the foundation model embedding space.
The cosine similarity baseline stores 4–5 fixed prototype embeddings per class extracted from the representative dataset, or sampled from the training set if none is available, and assigns the nearest class at inference time.
In this and the next experiment we use ground truth segmentation for evaluation.
The performance gain averaged over all city datasets is presented in Table \ref{tab:cosine_vs_hopfield_improvement}.

\begin{table}[t]
    \caption{Hopfield Heads with average performance gain over Cosine.}
    \label{tab:cosine_vs_hopfield_improvement}
    \begin{tabular}{lrrrrr}
        \toprule
         Model & mAP & Accuracy & Precision & Recall & F1-Score \\
        \midrule
        CLIP   & 0.699 (+0.178) & 0.570 (+0.136) & 0.720 (+0.121) & 0.720 (+0.121) & 0.720 (+0.121) \\
        Theia  & 0.704 (+0.056) & 0.565 (+0.037) & 0.716 (+0.029) & 0.716 (+0.029) & 0.716 (+0.029) \\
        ViT    & 0.701 (+0.058) & 0.559 (+0.054) & 0.710 (+0.043) & 0.710 (+0.043) & 0.710 (+0.043) \\
        \bottomrule
    \end{tabular}
    \vspace{-0.2cm}
\end{table}
Hopfield heads consistently outperform cosine similarity across all six city datasets and all three architectures, supporting \textbf{Q1}. CLIP benefits most, with average mAP rising from \textbf{0.521} to \textbf{0.699}. Theia reaches the highest individual average (\textbf{0.704 mAP}), yet the gap between models hints at complementary representations — motivation for the ensemble in Q2.

\textbf{Q2} For the second experiment we test each of the trained Hopfield Heads against their ensemble. Here, Theia Foundation Model is of interest as it is an embedding model distilled from CLIP, ViT, SAM, Depth Anything and DINOv2.
\begin{table}[t]
\centering
\caption{Ensemble improvement rate.}
\label{tab:ensemble_rate}
\begin{tabular}{lrrrrr}
\toprule
Model & mAP & Accuracy & Precision & Recall & F1-score \\
\midrule
CLIP & 0.694 & 0.565 & 0.715 & 0.715 & 0.715 \\
Theia & 0.699 & 0.552 & 0.705 & 0.705 & 0.705 \\
ViT & 0.704 & 0.555 & 0.709 & 0.709 & 0.709 \\
\textbf{Ensemble} & \textbf{0.821} & \textbf{0.703} & \textbf{0.824} & \textbf{0.824} & \textbf{0.824} \\
\bottomrule
\end{tabular}
    \vspace{-0.2cm}
\end{table}

We present the results in Table \ref{tab:ensemble_rate}. The empirical data clearly demonstrates the benefit of utilizing an Ensemble over the individual models. This supports our claim for \textbf{Q2} and shows that a single distilled representation may not always be an optimal solution to the problem of object recognition. The results also show that the chosen Foundation Models possess complementary representations and are a good fit for an Ensemble model.

\textbf{Q3} For the third experiment our approach is two-fold: we firstly evaluate the full pipeline Segment Proposer+Ensemble Head on the Ground Truth data, after which we evaluate it's efficiency in aiding the annotation process.

To set up the experiment, we divide the object classes into three categories determined by their shape: Simple - spherical and boxy objects (apple, cereal box), Medium - non-simple volumetric shapes (pack of chips, bottles, cups), Complex - stretched and precise shapes that take more variation to capture (Cutlery). We task five human experts to annotate a synthetic dataset and capture the difference in time necessary to correctly label the objects of given difficulty class. We present the result in Table \ref{tab:labeling_time}.
\begin{table}[t]
\centering
\caption{Average labeling time per complexity class.}
\label{tab:labeling_time}
\begin{tabular}{lll}
\toprule
Complexity & Per Dataset~~ & Per Object \\
\midrule
Simple (S) & 09:05.98 & 00:02.27 \\
Medium (M) & 09:46.22 & 00:02.44 \\
Complex (C) & 11:17.18 & 00:02.82 \\
\bottomrule
\end{tabular}
    \vspace{-0.2cm}
\end{table}


We evaluate our approach then, using the established shape categories of recognized objects. The results are available in Table \ref{tab:map_complexity}.

\begin{table}[t]
    \centering
    \caption{mAP by dataset, model, and object complexity.}
    \resizebox{0.9\textwidth}{!}{%
    \begin{tabular}{lrrrr | rrrr | rrrr | rrrr}
    \toprule
     & \multicolumn{4}{c}{CLIP} & \multicolumn{4}{c}{Theia} & \multicolumn{4}{c}{ViT} & \multicolumn{4}{c}{Ensemble} \\
    \cmidrule(lr){2-5} \cmidrule(lr){6-9} \cmidrule(lr){10-13} \cmidrule(lr){14-17}
    Dataset & S & M & C & \textbf{~All~} & S & M & C & \textbf{~All~} & S & M & C & \textbf{~All~} & S & M & C & \textbf{~All~} \\
    \midrule
    Bonn        & .65 & .71 & .21 & \textbf{.624} & ~.67 & .72 & .19 & \textbf{.637} & ~.65 & .73 & .23 & \textbf{.633} & ~.70 & .76 & .24 & \textbf{.671} \\
    Bordeaux    & .45 & .44 & .03 & \textbf{.349} & ~.43 & .44 & .03 & \textbf{.338} & ~.43 & .43 & .05 & \textbf{.337} & ~.53 & .54 & .05 & \textbf{.417} \\
    Eindhoven   & .55 & .48 & .39 & \textbf{.509} & ~.56 & .43 & .43 & \textbf{.505} & ~.56 & .43 & .41 & \textbf{.499} & ~.67 & .64 & .51 & \textbf{.638} \\
    Kassel      & .62 & .64 & .00 & \textbf{.607} & ~.61 & .66 & .00 & \textbf{.613} & ~.63 & .66 & .00 & \textbf{.622} & ~.71 & .73 & .00 & \textbf{.695} \\
    Cologne       & .44 & .51 & .19 & \textbf{.443} & ~.51 & .46 & .22 & \textbf{.451} & ~.54 & .54 & .27 & \textbf{.507} & ~.64 & .64 & .29 & \textbf{.601} \\
    Nuernberg   & .57 & .64 & .23 & \textbf{.558} & ~.60 & .60 & .20 & \textbf{.557} & ~.55 & .62 & .22 & \textbf{.538} & ~.70 & .73 & .33 & \textbf{.667} \\
    Salvador    & .63 & .59 & .33 & \textbf{.563} & ~.66 & .58 & .31 & \textbf{.569} & ~.62 & .57 & .34 & \textbf{.554} & ~.69 & .62 & .36 & \textbf{.604} \\
    \bottomrule
    \end{tabular}%
    }
    \label{tab:map_complexity}
    \vspace{-0.2cm}
\end{table}

The results confirm the ensemble benefit: mAP improves across all city datasets, with gains of up to 0.15 over individual heads (Eindhoven, Cologne). Even in Bordeaux, where complex objects are more varied, the ensemble maintains the lead. The final comparison averaged over city datasets is in Table~\ref{tab:map_complexity_final}.

\begin{table}[t]
    \centering
    \caption{Approach performance with Segment Proposer across datasets.}
    \begin{tabular}{lrrrrr}
    \toprule
    Model & Simple & Medium & Complex & All & Ground Truth \\
    \midrule
    CLIP & 0.560 & 0.572 & 0.199 & 0.522 & 0.713\\
    Theia & 0.579 & 0.557 & 0.198 & 0.524 & 0.716 \\
    ViT & 0.567 & 0.568 & 0.216 & 0.527 & 0.715 \\
    \textbf{Ensemble} & \textbf{0.663} & \textbf{0.664} & \textbf{0.255} & \textbf{0.613} & \textbf{0.825} \\
    \bottomrule
    \end{tabular}
    \label{tab:map_complexity_final}
    \vspace{-0.2cm}
\end{table}

The results also show the impact that the quality of Segment Proposer has on our approach, with mAP dropping by around 0.2 compared to running the Ensemble over the ground truth segmentation.

Finally, we utilize the result of the annotation time experiment (Table \ref{tab:labeling_time}) and the result of pipeline's accuracy (Table \ref{tab:map_complexity_final}) to estimate the time our approach saves human annotators on site. We present our finding in Table \ref{tab:objects_and_time_saved}.
\begin{table}[t]
    \caption{Objects retrieved and labeling time saved by the model.}
    \label{tab:objects_and_time_saved}
    \resizebox{\textwidth}{!}{%
    \begin{tabular}{lrrrrrrrr}
        \toprule
         & \multicolumn{3}{c}{Objects Retrieved (of GT)} & \multicolumn{4}{c}{Time Saved (of GT)}   & \%  \\
        \cmidrule(lr){2-4} \cmidrule(lr){5-8} 
        Dataset & Simple & Medium & Complex & Simple & Medium & Complex & Total & Saved\\
        \midrule
        Bonn      & 2850 (4065) & 2425 (3203) & 179 (732)   & 1:47:48 (2:33:47) & 1:38:36 (2:10:15) & 0:08:23 (0:34:24) & 3:34:48 (5:18:27) & 67.5\% \\
        Bordeaux  & 1315 (2471) & 916  (1703) & 86  (1714)  & 0:49:44 (1:33:29) & 0:37:15 (1:09:15) & 0:04:01 (1:20:33) & 1:31:01 (4:03:17) & 37.4\% \\
        Eindhoven & 2305 (3445) & 1095 (1721) & 173 (342)   & 1:27:11 (2:10:20) & 0:44:30 (1:09:59) & 0:08:08 (0:16:04) & 2:19:51 (3:36:23) & 64.6\% \\
        Kassel    & 1581 (2230) & 1356 (1858) & 0   (144)   & 0:59:49 (1:24:22) & 0:55:09 (1:15:33) & 0:00:00 (0:06:46) & 1:54:58 (2:46:41) & 69.0\% \\
        Cologne     & 1509 (2340) & 1335 (2082) & 73  (253)   & 0:57:06 (1:28:31) & 0:54:16 (1:24:40) & 0:03:26 (0:11:53) & 1:54:48 (3:05:05) & 62.0\% \\
        Nuernberg & 1484 (2120) & 1538 (2115) & 117 (353)   & 0:56:08 (1:20:12) & 1:02:31 (1:26:00) & 0:05:30 (0:16:35) & 2:04:10 (3:02:48) & 67.9\% \\
        Salvador  & 1992 (2900) & 1842 (2962) & 335 (929)   & 1:15:22 (1:49:43) & 1:14:55 (2:00:27) & 0:15:45 (0:43:39) & 2:46:03 (4:33:50) & 60.6\% \\
        \bottomrule
    \end{tabular}%
    }
    \vspace{-0.2cm}
\end{table}

The results empirically show the benefit of our approach and confirm that it is capable of taking over roughly 60\% of the data without correction. The only exception is Bordeaux dataset which has more variation in the objects of Complex category. This supports our claim for \textbf{Q3}.

\textbf{Q4} For the last experiment, we evaluate the quality of trained detectors. We compare MaskedDinoV2 models trained purely on the generated annotations to the ones that were post-corrected by human experts. We use both training and validation splits for the final evaluation and present our findings in Table \ref{tab:pipeline_vs_gt}.

\begin{table}[t]
\centering
\caption{Pipeline vs.\ Ground Truth performance by dataset.}
\label{tab:pipeline_vs_gt}
\resizebox{0.95\textwidth}{!}{
\begin{tabular}{llll|lll}
\toprule
 & \multicolumn{3}{c|}{Pipeline} & \multicolumn{3}{c}{Ground Truth} \\
 City & Precision & Recall & mAP & ~Precision & Recall & mAP \\
\midrule
Bonn      & .754 ±.004 & .835 ±.005 & .693 ±.004 & ~.914 ±.007 & .943 ±.004 & .797 ±.004 \\
Bordeaux  & .768 ±.007 & .584 ±.002 & .422 ±.002 & ~.867 ±.006 & .913 ±.001 & .698 ±.001 \\
Eindhoven & .788 ±.003 & .831 ±.005 & .671 ±.004 & ~.949 ±.005 & .955 ±.003 & .813 ±.002 \\
Kassel    & .668 ±.004 & .826 ±.004 & .703 ±.001 & ~.962 ±.003 & .968 ±.002 & .855 ±.007 \\
Cologne     & .725 ±.005 & .822 ±.002 & .646 ±.001 & ~.908 ±.002 & .944 ±.003 & .766 ±.002 \\
Nuernberg & .687 ±.009 & .818 ±.003 & .663 ±.003 & ~.869 ±.006 & .956 ±.004 & .811 ±.004 \\
Salvador  & .804 ±.006 & .837 ±.001 & .639 ±.002 & ~.927 ±.005 & .950 ±.003 & .776 ±.004 \\
\midrule
\textbf{Final} & \textbf{.742±.005} & \textbf{.793±.003} & \textbf{.634±.003} & \textbf{~.914±.005} & \textbf{.947±.003} & \textbf{.788±.003} \\
\bottomrule
\end{tabular}
}
\end{table}

The results indicate that while there is a 0.15 mAP loss in performance compared to expert-annotated models, they have decently good performance and can be used without human supervision, answering \textbf{Q4}.

\subsection{Competition Deployment}

\begin{figure}[t]
    \centering
    \scalebox{0.9}{
    \begin{tikzpicture}
        \node[anchor=south west, inner sep=0] (image1) at (0,0)
          {\includegraphics[height=0.17\textheight]{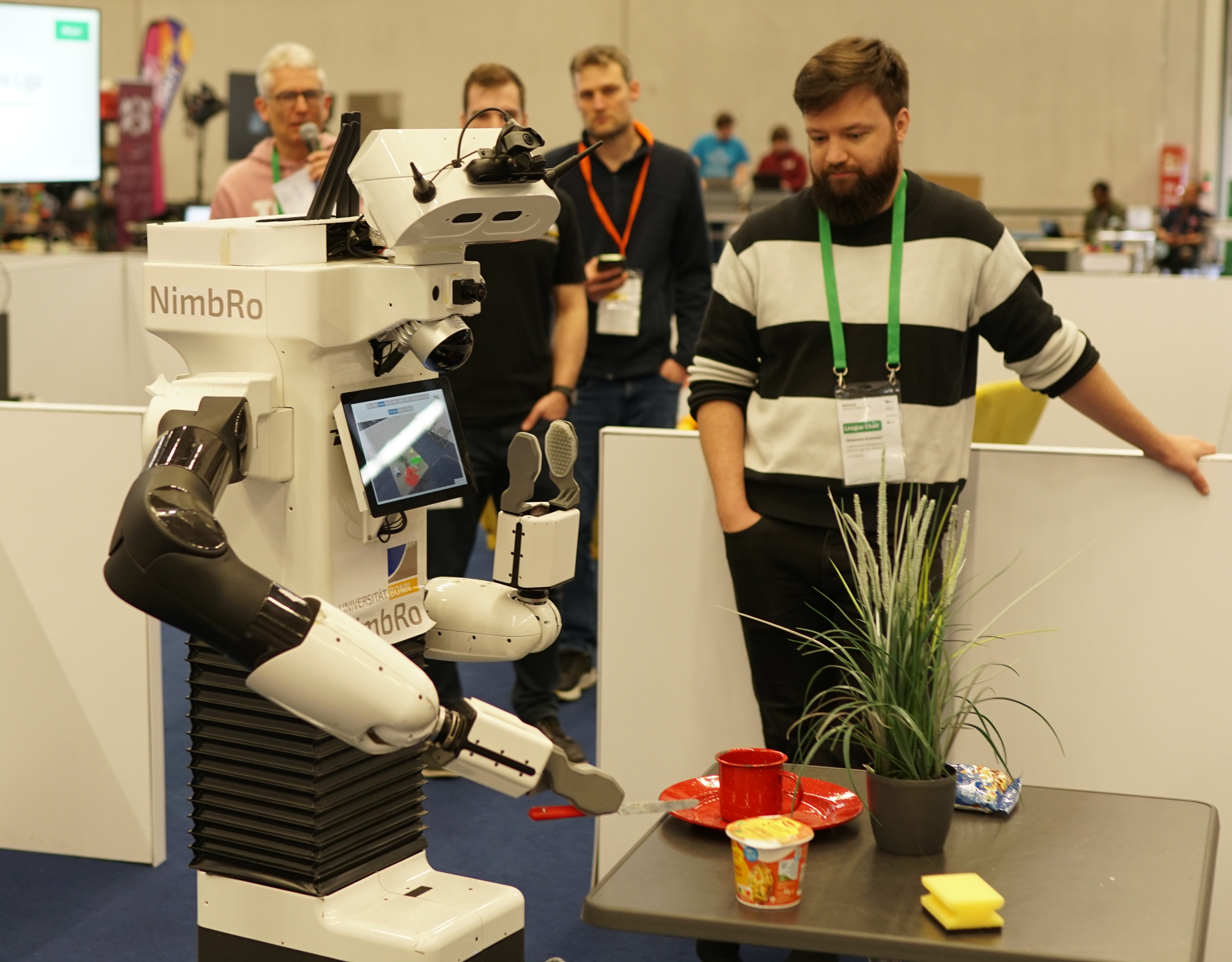}};
        \node[anchor=south west, inner sep=0] (image2) at (5,0)
          {\includegraphics[height=0.17\textheight]{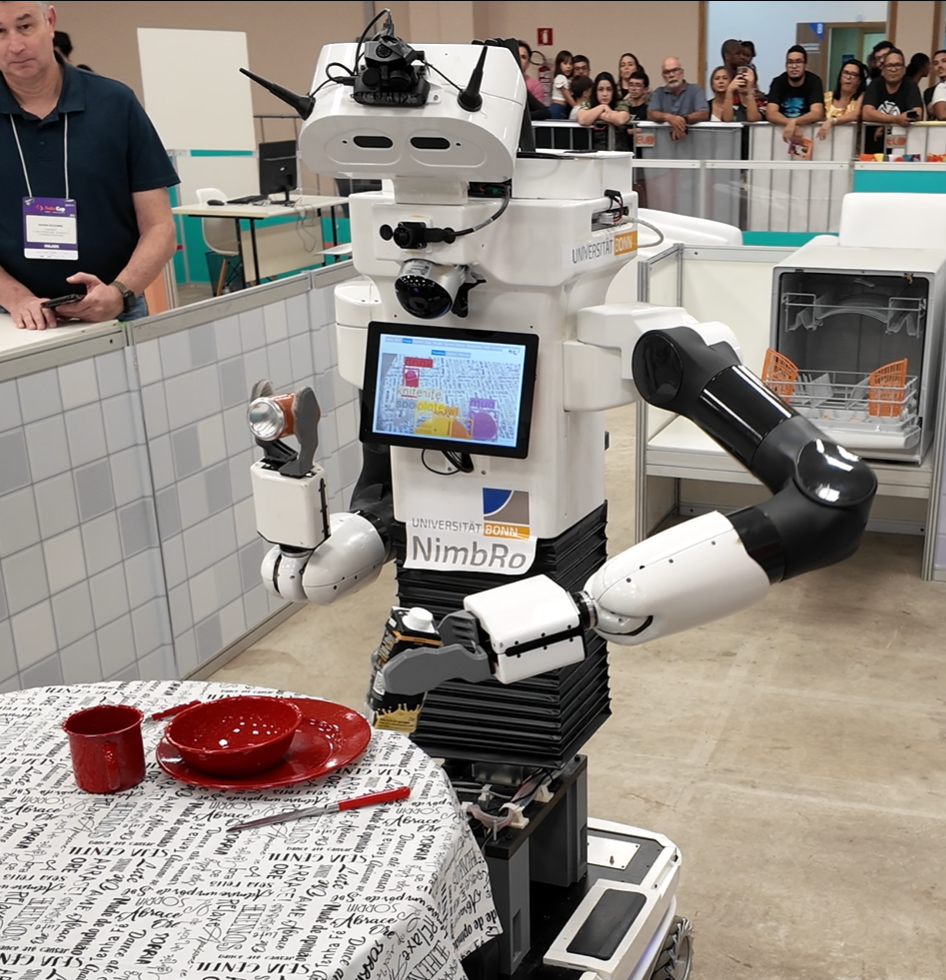}};
        \node[anchor=south west, inner sep=0] (image3) at (9,0)
          {\includegraphics[height=0.17\textheight]{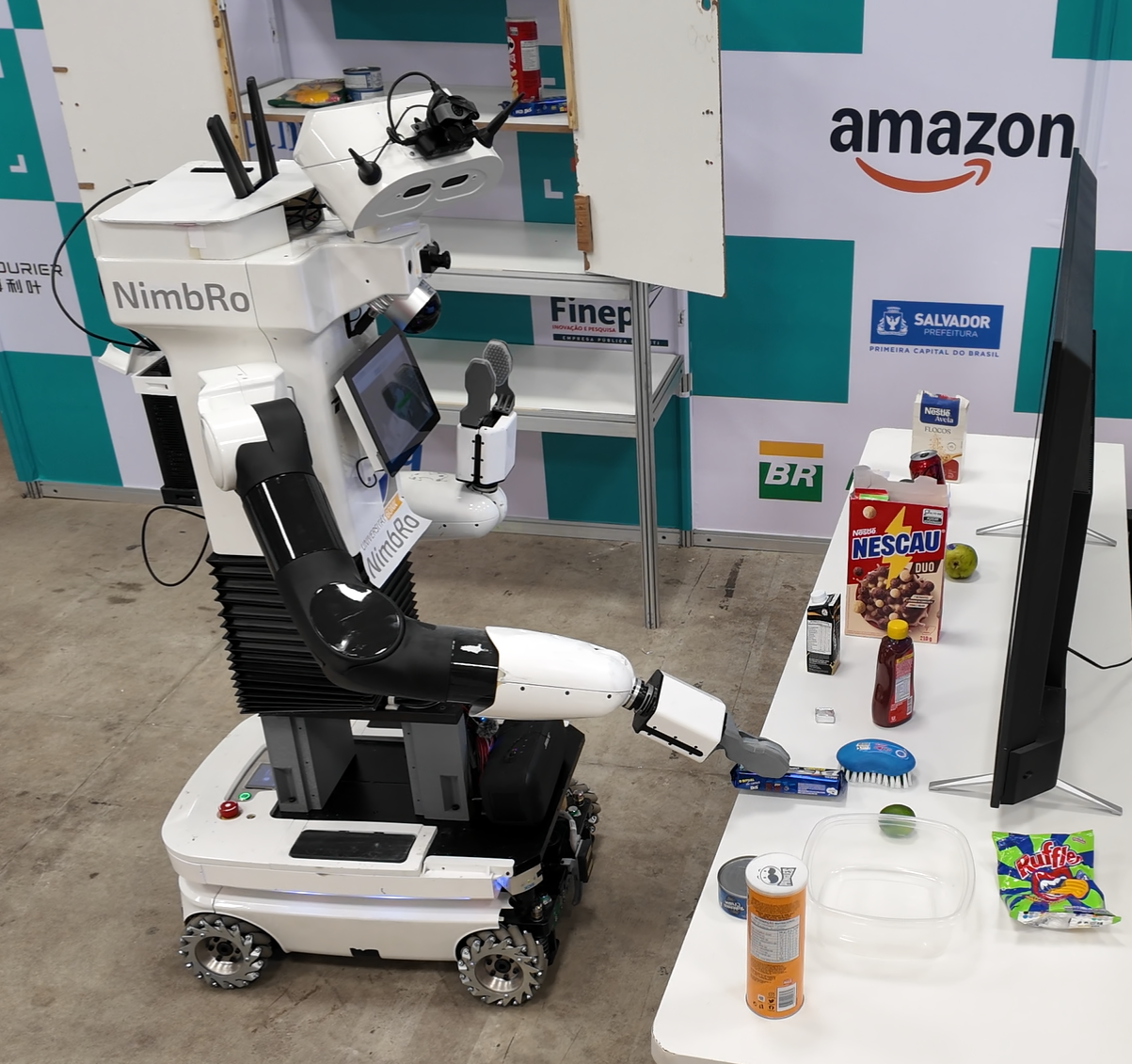}};
        \node[below=0.1cm of image1, font=\small] {Storing Groceries (RCGO 2026)};
        \node[below=0.1cm of image2, font=\small] {Clean the Table (RC 2025)};
        \node[below=0.1cm of image3, font=\small] {Storing Groceries (RC 2025)};
    \end{tikzpicture}
    }
    \vspace{-0.2cm}
    \caption{Examples of the tasks that the robot performed during the RoboCup@Home competitions. Our labeling approach allowed for efficient perception model training that resulted in successful grasping in multiple tasks.}
    \label{fig:robocup_examples}
    \vspace{-0.2cm}
\end{figure}

The presented approach has been deployed for the NimbRo@Home team at the RoboCup@Home world championship 2025 in Salvador, Brazil and the RoboCup @Home German Open 2026 in Cologne, Germany. Impressions of the participations are shown in \Cref{fig:robocup_examples}).
Task-specific segmentation models were trained using the pipeline, enabling on-site annotation and rapid model updates.
The approach is especially beneficial for manipulation-centric tasks such as Storing Groceries, General Purpose Service Robot, and Restaurant, where the robot must interact with many objects, and was successfully deployed for all of these.

\section{Conclusion}

In this paper, we presented a label propagation approach that estimates general object segmentations for household objects by applying a two-step approach: training a Segment Proposer to establish a shape bias and an ensemble of Hopfield networks to assign the label, trained only on a small representative dataset. We empirically show our approach to be more advantageous than simple metric comparison in Foundation Model embedding spaces. An ensemble of Hopfield heads effectively learned representative embeddings in complementary Foundation Model spaces and was successfully applied to both ground truth and generated by the Segment Proposer annotations. We successfully applied our approach on the gathered from previous RoboCup competitions datasets, which we make publicly available together with the annotated datasets. 

While the Labeler Heads take a small dataset to train, they must still be re-trained each time on the newly published at the competition objects. A direction for improvement would be then to introduce a "vocabulary" of the representatives that is extended on each next competition with previously unseen instances. This would reduce re-training overhead and alleviate the need for constructing a new representative dataset, beyond the instances not yet covered by the vocabulary.
Our experiments also show the importance of a reliable Segment Proposer as it contributes at least 0.2 mAP loss compared to ground truth segmentations. The future work could focus on experimenting with models that have better grounding and explore an additional filtering approach that would prevent out-of-distribution segments to be handed for recognition to the Ensemble model. 

Another promising next step is to extend our approach to video, which enables more efficient data capture, especially for RoboCup@Home. For instance, this may be done by using our method on a few frames and propagating the labels through remaining frames using tracking methods such as SAM-Video.

\section*{Acknowledgment}

The authors would like to thank Jan Nogga for his support in the data collection process and the preparation of the models.

 \printbibliography

\end{document}